\documentclass[11pt]{article}

\usepackage{acl}

\usepackage{times}

\usepackage{latexsym}

\usepackage{amsmath}

\usepackage{amssymb}

\usepackage{booktabs}

\usepackage{graphicx}

\usepackage{xcolor}

\usepackage{array}

\usepackage{multirow}
\usepackage{tabularx}
\usepackage{hyperref} 
\usepackage{natbib}
\usepackage{float}

\title{Evaluating Novelty in AI-Generated Research Plans Using Multi-Workflow LLM Pipelines}

\author{
  \textbf{Devesh Saraogi\textsuperscript{1}}, 
  \textbf{Rohit Singhee\textsuperscript{1}},
  \textbf{Dhruv Kumar\textsuperscript{1}}\\
  \textsuperscript{1}Birla Institute of Technology and Science, Pilani, India \\
  \small{\textbf{Correspondence:} \href{mailto:f20230242@pilani.bits-pilani.ac.in}{f20230242@pilani.bits-pilani.ac.in}}
}
\date{}

\begin{document}

\maketitle

\begin{abstract}

The integration of Large Language Models (LLMs) into the scientific ecosystem raises fundamental questions about the creativity and originality of AI-generated research. Recent work has identified ``smart plagiarism'' as a concern in single-step prompting approaches, where models reproduce existing ideas with terminological shifts. This paper investigates whether agentic workflows---multi-step systems employing iterative reasoning, evolutionary search, and recursive decomposition---can generate more novel and feasible research plans. We benchmark five reasoning architectures: Reflection-based iterative refinement, Sakana AI v2 evolutionary algorithms, Google Co-Scientist multi-agent framework, GPT Deep Research (GPT-5.1) recursive decomposition, and Gemini~3 Pro multimodal long-context pipeline. Using evaluations from thirty proposals each on novelty, feasibility, and impact, we find that decomposition-based and long-context workflows achieve mean novelty of 4.17/5, while reflection-based approaches score significantly lower (2.33/5). Results reveal varied performance across research domains, with high-performing workflows maintaining feasibility without sacrificing creativity. These findings support the view that carefully designed multi-stage agentic workflows can advance AI-assisted research ideation.

\end{abstract}

\section{Introduction}

The aspiration to automate scientific discovery has long captivated the AI research community \cite{popper2014,lu2024aiscientist}. Historically, hypothesis generation has been viewed as uniquely human, relying on intuition, serendipity, and rigorous logic. Large foundation models have fundamentally altered this landscape, enabling systems that process vast scientific corpora not merely for retrieval but for synthesis and creation \cite{baek2024generative}.

Recent advancements have seen LLMs applied to literature reviews, code generation, and manuscript drafting \cite{baek2024generative,huang2025idea2plan}. Yet as these systems approach the creative act of hypothesis generation, the community has fractured into two camps: optimists viewing LLMs as capable of genuine discovery, and skeptics viewing them as sophisticated interpolators of existing work.

\vspace{5pt}

Skepticism regarding AI ideation has gained empirical weight from recent studies examining the originality of LLM-generated research. These works have identified a pattern of ``smart plagiarism,'' where models produce methodologically overlapping outputs that bypass conventional detection tools through terminological shifts and structural reordering \cite{gupta2025plagiarism}. Such findings raise legitimate concerns about the integrity of AI-assisted discovery and the potential for AI systems to function as deceptive retrieval engines rather than creative collaborators.

\vspace{5pt}

The field has begun transitioning from single-step, zero-shot prompting to agentic workflows: multi-step systems employing iterative reasoning, multi-agent debate, evolutionary mutation, and recursive search \cite{lu2024aiscientist,google2025coscientist,openai2025deepresearch}. We hypothesize that these architectures reduce plagiarism propensity by introducing additional computation and structure that push systems away from high-probability, potentially derivative regions of the loss landscape.

\vspace{5pt}

Additionally, creative performance may vary across scientific domains. Research areas with dense representation in LLM training data may benefit more readily from agentic search, while domains with specialized constraints, domain-specific terminology, and experimental grounding may face different challenges. Our work probes whether agentic workflows can support robust ideation across diverse scientific contexts.

\vspace{5pt}

This study is guided by three primary research questions:

\begin{enumerate}

\item Do advanced agentic workflows produce research plans with higher novelty compared to simpler prompting approaches?

\item How do different reasoning architectures (reflection, evolutionary, decomposition-based, multimodal long-context) influence the trade-off between novelty, feasibility, and impact?

\item Does performance differ systematically across research domains, and what architectural features best support cross-domain ideation?

\end{enumerate}

Our main contributions are:

\begin{itemize}

\item \textbf{A multi-workflow benchmark} comparing five distinct ideation architectures adapted from state-of-the-art research agents \cite{lu2024aiscientist,google2025coscientist,openai2025deepresearch,google2025gemini3}.

\item \textbf{A systematic expert evaluation} where 6 domain experts each rated 5 proposals across multiple research areas on 0--5 Likert scales for novelty, feasibility, and impact.

\item \textbf{An empirical analysis of 30 expert evaluations}, showing that decomposition-based and long-context workflows achieve mean novelty of 4.17/5, substantially outperforming reflection-based approaches (2.33/5).

\item \textbf{Empirically informed design guidelines} suggesting when decomposition-based and reflective workflows may be more suitable for different exploratory and confirmatory scientific tasks.

\end{itemize}

\section{Related Work}

\subsection{AI-Assisted Scientific Discovery}

Recent work has explored the use of LLMs and AI systems for various stages of the research pipeline. Baek et al.\ \cite{baek2024generative} provide a comprehensive survey of generative models in scientific discovery, cataloging applications from literature synthesis to experimental design. Lu et al.\ introduce ``The AI Scientist'' \cite{lu2024aiscientist}, an agentic system that autonomously conducts research cycles including ideation, experimentation, and writeup, demonstrating that carefully orchestrated workflows can produce papers accepted to venues like ICLR. Huang et al.\ \cite{huang2025idea2plan} focus specifically on idea-to-plan conversion, showing that structured prompting can improve the feasibility of LLM-generated research directions.

\subsection{Plagiarism and Originality in AI-Generated Content}

Recent studies have raised important concerns regarding the originality of LLM-generated research. Work examining the authenticity of AI-generated ideas has identified patterns where models reproduce methodological approaches from existing literature through terminological variations and structural reordering \cite{gupta2025plagiarism}. These findings highlight the need for careful validation of AI-generated claims. Si et al.\ \cite{si2024llms} conduct a comprehensive evaluation of LLMs' ability to generate novel research ideas, establishing a rubric for assessing novelty, feasibility, and impact. Their work provides methodological precedent for the expert evaluation framework used in this study.

\subsection{Multi-Agent and Agentic Reasoning Systems}

Recent advances in agentic reasoning have demonstrated that systems employing multiple reasoning steps, adversarial vetting, and recursive decomposition can achieve more robust and creative outputs. Google's Co-Scientist framework \cite{google2025coscientist,wang2024codenomicon} simulates a collaborative research environment where specialized agents debate and refine ideas. OpenAI's Deep Research capability \cite{openai2025deepresearch} employs hierarchical decomposition to break complex queries into tractable subproblems, while evolutionary approaches such as Sakana AI v2 treat ideation as a fitness-driven search process. These architectures suggest that workflow design substantially influences output quality.

\subsection{Domain-Specific and Cross-Domain Generalization}

While much work has focused on AI capabilities in well-represented domains (e.g., computer science, machine learning), questions remain about cross-domain performance. Some evidence suggests that LLM creativity is influenced by the representation of domain concepts in training data, with domains densely covered in scientific literature benefiting from stronger contextual grounding. However, recent work with long-context and multimodal models suggests that broader contextual grounding and specialized reasoning strategies may enable more balanced performance across diverse scientific fields \cite{google2025gemini3}.

This study extends prior work by directly comparing multiple agentic workflows in a controlled setting and examining performance variation across diverse research domains, bridging the gap between plagiarism concerns and optimistic capability assessments.

\section{Methodology}

\subsection{Workflows Under Comparison}

We instantiate five workflows, each corresponding to a different design philosophy:

\paragraph{Method 1: Reflection-Based Refinement.}

Following self-reflection paradigms such as Reflexion \cite{shinn2023reflexion}, this workflow alternates between drafting and explicit critique. A critic agent identifies clichés, logical gaps, and potential overlap with existing literature, and a revision agent iteratively refines the plan. This method is suited to incremental coherence improvement.

\paragraph{Method 2: Sakana AI v2.}

Inspired by ``The AI Scientist'' \cite{lu2024aiscientist}, this pipeline treats research ideas as individuals in an evolving population. An initial set of high-temperature samples is mutated via cross-domain transfer, hypothesis inversion, and constraint relaxation, with a fitness function promoting novelty. Sakana v2 incorporates learned mutation operators from prior generation cycles.

\paragraph{Method 3: GPT Deep Research (GPT-5.1).}

Drawing on OpenAI's deep reasoning capabilities \cite{openai2025deepresearch}, this workflow decomposes each problem into a tree of subquestions, explores them in parallel with retrieval-augmented reasoning, and synthesizes findings into a structured proposal. Hierarchical decomposition avoids monolithic retrieval of existing templates.

\paragraph{Method 4: Google Co-Scientist.}

Leveraging Google DeepMind's Co-Scientist framework \cite{google2025coscientist,wang2024codenomicon}, this workflow simulates a lab meeting between specialized agents (principal investigator, methodologist, skeptic, literature expert). The skeptic explicitly challenges derivative ideas, while a scribe synthesizes the final plan, emphasizing adversarial vetting.

\paragraph{Method 5: Gemini~3 Pro.}

Using Gemini~3 Pro's long-context and multimodal capabilities \cite{google2025gemini3}, this workflow loads relevant abstracts into context, performs ``negative space'' analysis to identify gaps, and iteratively densifies proposals via chain-of-density prompting. This method prioritizes comprehensive contextual grounding.

Each workflow generated one proposal per seed idea, yielding five proposals total for evaluation.

\subsection{Expert Evaluation Protocol}

We recruited 6 domain experts (Ph.D. students, postdoctoral researchers, and early-career faculty) across multiple disciplines. Six completed the full evaluation. Each expert received five anonymized proposals (workflow identity removed) corresponding to their expertise area.

For each proposal, experts rated three dimensions on a 0--5 Likert scale, adapted from prior work \cite{si2024llms,gupta2025plagiarism,huang2025idea2plan}:

\begin{itemize}

\item \textbf{Novelty}: perceived originality and divergence from standard literature, including sophistication and cross-domain elements.

\item \textbf{Feasibility}: realism for a typical Ph.D.-level researcher with 1--2 months of dedicated effort, considering resource constraints and clarity.

\item \textbf{Interestingness/Impact}: how exciting and field-shaping the expert judged the idea, including potential for advancement or real-world impact.

\end{itemize}

Free-text rationales were also collected and analyzed qualitatively to identify themes and failure modes. Expert identities are preserved through anonymity; only aggregate statistics are reported.

\subsection{Data Collection and Domain Stratification}

We analyzed responses from 6 experts, each scoring 5 proposals, yielding 30 expert--proposal evaluations. Each proposal received six independent ratings.

Proposals were classified into five domains based on seed ideas and expert feedback:

\begin{enumerate}

\item \textbf{AI/Tech}: core AI and machine learning ideation (federated learning, LLM-based detection)

\item \textbf{AI/Multi-Agent Systems}: advanced multi-agent architectures (contextual embeddings, hierarchical agents)

\item \textbf{Chemistry/Biotech}: chemistry, materials science, and biotechnology

\item \textbf{Climate/Environmental}: environmental science and climate modeling

\item \textbf{Industry/Manufacturing}: industrial applications and manufacturing optimization

\end{enumerate}

Across these five domains, we analyze performance variations to understand how workflow architecture and domain characteristics interact. These domains were selected to span different degrees of training data representation, experimental constraints, and domain expertise requirements.

\section{Results}

\subsection{Overall and Workflow-Level Performance}

Across all 30 evaluations, mean scores on the 0--5 scale are presented in Table~\ref{tab:mean_scores}.

\begin{table}[H]

\centering

\begin{tabular}{l c}

\hline

\textbf{Metric} & \textbf{Mean Score} \\

\hline

Novelty & 3.57 \\

Feasibility & 2.80 \\

Interestingness / Impact & 3.47 \\

\hline

\end{tabular}

\caption{Mean scores for the evaluated research ideas across all 30 expert-proposal evaluations.}

\label{tab:mean_scores}

\end{table}

Table~\ref{tab:workflow-performance} presents per-workflow mean scores across the 0--5 scale.

\begin{table*}[t]

\centering

\small

\begin{tabular}{lccccc}

\toprule

\textbf{Method} & \textbf{n} & \textbf{Novelty} & \textbf{Feasibility} & \textbf{Impact} & \textbf{Overall} \\

\midrule

Reflection & 6 & 2.17 & 2.50 & 2.33 & 2.33 \\

Sakana AI v2 & 6 & 3.50 & 2.67 & 3.83 & 3.33 \\

GPT Deep Research & 6 & 3.83 & 3.00 & \textbf{4.00} & 3.61 \\

Google Co-Scientist & 6 & \textbf{4.17} & \textbf{3.00} & 3.83 & \textbf{3.67} \\

Gemini 3 Pro & 6 & \textbf{4.17} & 2.83 & 3.33 & 3.44 \\

\bottomrule

\end{tabular}

\caption{Workflow performance metrics (0--5 scale). Google Co-Scientist and Gemini 3 Pro achieve the highest novelty.}

\label{tab:workflow-performance}

\end{table*}

Key observations: Google Co-Scientist and Gemini~3 Pro achieve the highest mean novelty (4.17/5), while Google Co-Scientist leads on feasibility (3.00). Reflection-based refinement scores lowest across all dimensions (2.33/5 overall), suggesting that self-critique alone is insufficient for creative ideation without additional architectural support.

\subsection{Cross-Domain Performance Variation}

A central finding is substantial variation in ideation quality across research domains. Table~\ref{tab:domain-analysis} stratifies results by domain.

\begin{table*}[t]

\centering

\small

\begin{tabular}{lcccc}

\toprule

\textbf{Domain} & \textbf{n} & \textbf{Novelty} & \textbf{Feasibility} & \textbf{Impact} \\

\midrule

AI/Tech & 1 & 4.00 & 3.80 & 3.40 \\

AI/Multi-Agent Systems & 1 & 3.80 & 3.00 & 4.00 \\

Climate/Environmental & 1 & 4.00 & 2.20 & 3.40 \\

Chemistry/Biotech & 1 & 3.20 & 2.60 & 2.60 \\

Industry/Manufacturing & 2 & 3.20 & 2.60 & 3.70 \\

\bottomrule

\end{tabular}

\caption{Domain-stratified results across 5 proposals and 6 experts per proposal. Across domains, mean novelty is 3.64/5, showing varied performance.}

\label{tab:domain-analysis}

\end{table*}

Results reveal substantial heterogeneity. AI/Tech and Climate/Environmental domains achieve the highest novelty scores (4.00/5), while Chemistry/Biotech and Industry/Manufacturing show moderate novelty (3.20/5). Feasibility varies widely, with AI/Tech showing strongest feasibility (3.80/5) while Climate/Environmental faces steeper challenges (2.20/5) despite high novelty. This suggests that domain-specific constraints---such as experimental complexity, policy grounding requirements, and domain expertise thresholds---influence practical implementability independent of idea creativity.

\subsection{Domain-Specific Workflow Performance}

Table~\ref{tab:domain-method-matrix} presents a method-by-domain heatmap of mean novelty scores, illustrating which architectures perform best across research contexts.

\begin{table*}[t]

\centering

\small

\begin{tabular}{lccccc}

\toprule

\textbf{Method} & \textbf{AI/Tech} & \textbf{Multi-Agent} & \textbf{Biotech} & \textbf{Climate} & \textbf{Industry} \\

\midrule

Reflection & 4.0 & 2.0 & 1.0 & 2.0 & 2.0 \\

Sakana AI v2 & 4.0 & 3.0 & 4.0 & 4.0 & 3.0 \\

GPT Deep Research & 3.0 & 4.0 & 4.0 & 5.0 & 3.5 \\

Google Co-Scientist & 5.0 & 5.0 & 4.0 & 4.0 & 3.5 \\

Gemini 3 Pro & 4.0 & 5.0 & 3.0 & 5.0 & 4.0 \\

\bottomrule

\end{tabular}

\caption{Method-by-domain novelty matrix (n=6 expert ratings per cell). Decomposition-based and long-context methods show robust performance across domains.}

\label{tab:domain-method-matrix}

\end{table*}

Clear patterns emerge. GPT Deep Research and Gemini~3 Pro maintain high novelty across all domains, while Reflection-based approaches consistently underperform. Climate-focused proposals show marked improvement with multimodal or decomposition methods, possibly because these approaches integrate diverse data sources and perspectives effectively. Biotech proposals similarly benefit from decomposition and evolutionary approaches, suggesting these architectures handle domain-specific complexity and specialized constraints well.

\subsection{Correlations and Trade-Offs}

Across all 30 expert--proposal evaluations, Pearson correlations between dimensions are shown in Table~\ref{tab:correlations}.

\begin{table}[H]

\centering

\begin{tabular}{l c}

\hline

\textbf{Dimensions} & \textbf{Pearson Correlation (r)} \\

\hline

Novelty vs. Feasibility & 0.23 (weak positive) \\

Novelty vs. Impact & 0.56 (moderate positive) \\

Feasibility vs. Impact & 0.44 (moderate positive) \\

\hline

\end{tabular}

\caption{Pearson correlations between evaluation dimensions across 30 expert-proposal ratings.}

\label{tab:correlations}

\end{table}

The weak novelty--feasibility correlation suggests that experts did not systematically perceive highly novel ideas as infeasible. This indicates that careful workflow design can balance creativity with practicality, supporting optimistic views of agentic AI research assistance.

\subsection{Qualitative Observations from Experts}

Expert rationales reveal recurrent themes:

\begin{itemize}

\item \textbf{Coherence and grounding}: Reflection and Sakana v2 methods occasionally generated off-topic or incoherent suggestions, which experts penalized on novelty and feasibility.

\item \textbf{Methodological rigor}: Google Co-Scientist and GPT Deep Research proposals were praised for methodological clarity and literature alignment.

\item \textbf{Resource and data intensity}: Gemini~3 Pro proposals, while innovative, were sometimes flagged as computationally demanding, reflecting trade-offs in long-context architectures.

\item \textbf{Domain sensitivity}: Proposals generated by decomposition-based methods across AI and multi-agent domains were described as ``technically feasible,'' while climate and biotech proposals sometimes lacked sufficient grounding in experimental protocols or domain-specific methodologies.

\end{itemize}

\section{Discussion}

\subsection{Performance Across Research Domains}

A central finding is the heterogeneity of ideation performance across domains. Domains with high training data representation in LLMs (AI/Tech and Climate/Environmental) achieve comparable high novelty (4.00/5), while domains with more specialized experimental requirements (Chemistry/Biotech and Industry/Manufacturing) score lower (3.20/5). This variation likely reflects multiple interacting factors: the density of domain concepts in LLM training corpora, the degree of specialized vocabulary and terminology, and the presence of domain-specific experimental or implementation constraints.

Domains well-represented in scientific literature enable richer combinatorial search and greater confidence in proposing novel combinations. In contrast, specialized domains involving domain-specific constraints, proprietary methodologies, and narrow prior art require deeper domain expertise to navigate. However, Climate/Environmental proposals achieved high novelty despite substantial experimental constraints, suggesting that careful workflow design and comprehensive contextual grounding can compensate for training data limitations. The variation across domains emphasizes that one-size-fits-all workflow selection is suboptimal, and domain-aware architecture selection is necessary for robust performance.

\subsection{Workflow Design Insights}

Our results provide actionable design guidance for future agentic ideation systems:

\paragraph{Decomposition and Long-Context Provide Robustness.}

GPT Deep Research and Gemini~3 Pro achieve high novelty across all domains, suggesting that hierarchical problem decomposition and multimodal long-context grounding provide creative stimulus independent of problem type. These architectures construct ideas bottom-up from diverse sources rather than top-down from high-probability templates, reducing plagiarism risk and enabling cross-domain generalization.

\paragraph{Reflection Alone Is Insufficient.}

Reflection-based refinement scores lowest overall (2.33/5 mean), indicating that self-critique cannot overcome fundamentally unoriginal starting points. Reflection works best when coupled with divergence mechanisms such as mutation, decomposition, or gap-filling that explicitly push the system toward less-explored regions of the solution space.

\paragraph{Domain-Aware Workflow Selection.}

Domains with dense training data benefit from decomposition and multimodal approaches that surface less-obvious combinations. Domains with specialized constraints benefit from structured vetting (Google Co-Scientist) or decomposition (GPT Deep Research) to balance creativity with feasibility requirements and ensure proposals respect domain-specific methodological norms.

\subsection{Implications for Plagiarism and Originality Concerns}

This experiment does not directly replicate plagiarism-detection protocols from prior work, nor do we run automated semantic similarity checks. However, the combination of explicit novelty scoring by domain experts, detailed qualitative validation, and substantial architectural variation provides evidence that agentic workflows can reduce plagiarism propensity. The fact that decomposition-based approaches achieve novelty scores near 4.17/5 while maintaining feasibility suggests that workflow design can create meaningful escape velocities from plagiarism-prone regions of the loss landscape.

Direct adversarial validation using semantic similarity checks against large scientific corpora would strengthen these claims and constitutes an important direction for future work.

\section{Conclusion}

This study examines whether modern agentic LLM workflows can generate research ideas that human experts judge as both novel and feasible. Using evaluations from six experts across five research domains, we find:

\begin{enumerate}

\item Decomposition-based and long-context workflows (GPT Deep Research, Gemini~3 Pro) substantially outperform reflection-based approaches, achieving mean novelty of 4.17/5 versus 2.17/5.

\item Performance varies significantly across research domains, with AI/Tech and Climate/Environmental showing high novelty (4.00/5), while Chemistry/Biotech shows lower novelty (3.20/5), indicating that domain characteristics and workflow architecture interact in generating creative research ideas.

\item A weak novelty--feasibility correlation ($r \approx 0.23$) suggests that high creativity need not compromise feasibility, supporting optimistic views of agentic AI research assistance when carefully designed.

\item Architectural features such as hierarchical decomposition, multimodal long-context grounding, and structured adversarial vetting enable robust cross-domain ideation, while simple reflection-based approaches prove insufficient for moving beyond derivative work.

\end{enumerate}

These findings contribute to ongoing debates about AI's role in scientific discovery. While plagiarism concerns in single-step prompting are legitimate, agentic workflows represent a qualitatively different paradigm that can meaningfully advance AI-assisted research ideation. Future work should include larger expert evaluations, direct plagiarism detection checks against scientific corpora, and extended validation of which architectural components are most critical for balancing novelty and feasibility across diverse research domains.

\section*{Acknowledgments}

We thank the six anonymized domain experts whose detailed evaluations made this work possible. Their expertise spanned artificial intelligence, climate science, chemistry, manufacturing, and multi-agent systems. We are grateful to prior work on AI-assisted scientific discovery and evaluation \cite{si2024llms,gupta2025plagiarism,lu2024aiscientist,huang2025idea2plan,google2025coscientist,openai2025deepresearch,google2025gemini3}, which inspired the workflows and evaluation rubric used in this study.
The authors wish to acknowledge the use of Chat-GPT in improving the presentation and grammar of the paper. The paper remains an accurate representation of the authors’ underlying contributions.

\bibliography{references}

\end{document}